\documentclass[10pt,twocolumn,letterpaper]{article}

\usepackage[pagenumbers]{cvpr} % To force page numbers, e.g. for an arXiv version

\usepackage{graphicx}
\usepackage{amsmath}
\usepackage{amssymb}
\usepackage{booktabs}

\usepackage[pagebackref,breaklinks,colorlinks]{hyperref}

\usepackage[capitalize]{cleveref}
\crefname{section}{Sec.}{Secs.}
\Crefname{section}{Section}{Sections}
\Crefname{table}{Table}{Tables}
\crefname{table}{Tab.}{Tabs.}

\def\cvprPaperID{*****} % *** Enter the CVPR Paper ID here
\def\confName{CVPR}
\def\confYear{2022}

\begin{document}

\title{Single-image Defocus Deblurring by Integration of Defocus Map Prediction Tracing the Inverse Problem Computation}

\author{
    % Authors
    Qian Ye\textsuperscript{\rm 1},
    Masanori Suganuma\textsuperscript{\rm 1,2},
    % Jun Xiao\textsuperscript{\rm 3},
    Takayuki Okatani\textsuperscript{\rm 1,2}\\
    \textsuperscript{\rm 1} Graduate School of Information Sciences, Tohoku University \\
    \textsuperscript{\rm 2} RIKEN Center for AIP \\
    % \textsuperscript{\rm 3} Department of Electronic and Information Engineering, The Hong Kong Polytechnic University \\
    \{qian, suganuma, okatani\}@vision.is.tohoku.ac.jp 
}

% \and
\maketitle

%%%%%%%%% ABSTRACT
\begin{abstract}
In this paper, we consider the problem in defocus image deblurring. Previous classical methods follow two-steps approaches, i.e., first defocus map estimation and then the non-blind deblurring. In the era of deep learning, some researchers have tried to address these two problems by CNN. However, the simple concatenation of defocus map, which represents the blur level, leads to suboptimal performance. Considering the spatial variant property of the defocus blur and the blur level indicated in the defocus map, we employ the defocus map as conditional guidance to adjust the features from the input blurring images instead of simple concatenation. Then we propose a simple but effective network with spatial modulation based on the defocus map. To achieve this, we design a network consisting of three sub-networks, including the defocus map estimation network, a condition network that encodes the defocus map into condition features, and the defocus deblurring network that performs spatially dynamic modulation based on the condition features. Moreover, the spatially dynamic modulation is based on an affine transform function to adjust the features from the input blurry images. Experimental results show that our method can achieve better quantitative and qualitative evaluation performance than the existing state-of-the-art methods on the commonly used public test datasets. 
\end{abstract}

%%%%%%%%% BODY TEXT
\section{Introduction}
\label{sec:intro}

% Please follow the steps outlined below when submitting your manuscript to the IEEE Computer Society Press.
% \section{Introduction}

Defocus blur is inevitable when the rays from a scene not lying on the focal plane of the camera converge to a region rather than a point on the image plane and the region is called the circle of confusion (COC) \cite{potmesil1982synthetic}. Using a large aperture allows more light to pass through the lens in a shorter exposure time. But this results in a shallow depth of field (DoF), thereby causing defocus blur. Shallow DOF is useful to make the subject stand out from the blurry
background and foreground. But on the other hand, defocus blur causes visual information losses, which is important for other tasks like image understanding. Thus, recovering the blurry images can help improve the performance on these tasks. 

However, defocus deblurring is still a challenging task due to the spatial variant blur, i.e., the level of blur for each pixel is different. For example, the scene on the focal plane is captured sharply while the scene out of the focal plane is captured in blurry. And the level of defocus is usually depicted by the point spread function (PSF), resulting in a pixel-wise defocus map. 

The defocus blur can be modeled by 
\begin{equation}
I_b = K*I_c + N
\end{equation}
\noindent where $I_b$ is the blurry image, $I_c$ is the clean image, $K$ is the blur kernel and $N$ is the additive noise \cite{campisi2017blind}.  
Thus the natural idea for defocus deblurring is to follow two steps to address this problem, i.e., first to estimate the defocus map indicating the level of defocus blur, then apply non-blind deconvolution\cite{tai2009single,zhuo2011defocus,karaali2017edge}. These classical methods can achieve satisfying results under the low or medium level defocus blur but it is hard for these methods to achieve sharp results under high level blur. Some image priors are utilized to improve the performance but these priors works well under some particular scenes but may fail to cover the real-world scenes. What is more, the classical methods need iterative optimization to achieve sharp results, which is time-consuming.  

In the era of deep learning, many CNN-based based approaches are proposed for deblurring problem. Nah et al. \cite{nah2017deep} propose multi-scale structure for dynamic scene deblurring and achieve good visual results. Kupyn et al. \cite{kupyn2018deblurgan,kupyn2019deblurgan} utilize the generative adversarial networks (GANs) \cite{goodfellow2014generative} for the deblurring. Cho et al. \cite{cho2021rethinking} propose a multi-input
multi-output U-net (MIMO-UNet) to improve the performance while reduce the computational cost. Zamir et al. \cite{zamir2021multi} propose a multi-stage and multi-scale architecture for motion blur removal.   
And Chen et al. \cite{chen2022simple} propose a baseline model that achieves better performance and lower inference time compared with previous methods. But the above methods focus on the motion blur and therefore may be unsuitable for the defocus deblurring. 

Recently, there are also some CNN-based methods proposed for defoucs blurring \cite{abuolaim2020defocus,lee2021iterative,son2021single}. Abuolaim et al. \cite{abuolaim2020defocus} use the UNet structure to recover the sharp image in an end-to-end manner. Lee et al. \cite{lee2021iterative} propose an iterative filter adaptive module to handle spatially-varying and large defocus blur and a training scheme based on defocus disparity estimation and re-blurring to boost the deblurring quality. Son et al. \cite{son2021single} propose to simulate inverse kernels by kernel sharing parallel atrous convolution block.

However, there still is a CNN-based method that follows first defocus map estimation and then non-blind deblurring \cite{ma2021defocus}. Compared with previous CNN-based methods, the two-steps CNN-based methods can achieve higher performance since the bind deblurring problem is much more complicated than the non-blind deblurring problem, while the two steps CNN-based methods can utilize the important information from the defocus map. After the estimation of the defocus map using convolutional neural network, Ma et al. \cite{ma2021defocus} concatenates the defocus map and the defocus image as input of the neural network to achieve all sharp images. We argue that direct concatenation can not fully use the information in defocus map, which represents the blur level.  

Considering the spatial variant property of the defocus blur and the blur level indicated in the defocus map, we employ the defocus map as conditional guidance to adjust the features from the input blurring images instead of simple concatenation. Then we propose a simple but effective network with spatial modulation based on the defocus map.
To achieve this, we design a network consisting of three sub-networks, including the defocus map estimation network, a condition network that encodes the defocus map into condition features and the defocus deblurring network that performs spatially dynamic modulation based on the condition features. And the spatially dynamic modulation is achieved by an affine transform function to adjust the features from the input blurry images. Experimental results show that our method can achieve better quantitative and qualitative evaluation performance than the existing state-of-the-art methods on the commonly used public test datasets. 

%------------------------------------------------------------------------- 

\section{Related Work}

In this section, we briefly introduce the related works, including defocus map estimation, non-blind deblurring, and single image defocus deblurring.

\subsection{Defocus Map Estimation}

There are different approaches to defocus map estimation, which can be roughly divided into three categories: edge-based, region-based, and CNN-based methods. 

The basic idea of the edge-based methods is to compute a sparse defocus map only at the edges of the images and then propagate the map to the whole image \cite{pentland1987new,elder1998local,bae2007defocus,zhuo2011defocus,liu2016defocus}. In \cite{pentland1987new}, the input image is re-blurred by the Gaussian kernels, and then the defocus amount along the edges is computed by the rate of the gradients of the re-blurred image. Similar to \cite{pentland1987new}, Elder and Zucker \cite{elder1998local} propose a method that simultaneously detects image edges and estimates blur. These two studies \cite{pentland1987new,elder1998local} only focus on the sparse defocus map estimation. Liu et al. \cite{liu2016defocus} propose a two-parameter model to improve the performance of the defocus map estimation on the edges. Bae and Durand \cite{bae2007defocus} try to estimate the defocus map on the whole image. After the sparse defocus map estimation, their method uses a bilateral filter to remove outliers. It then uses a colorization-scheme-based interpolation method to achieve the full defocus map. Zhuo and Sim \cite{zhuo2011defocus} propose using alpha Laplacian Matting to propagate the sparse defocus map to the whole image. These methods suffer from inaccurate defocus map estimation for the image areas far from the edges.
In \cite{karaali2017edge}, a connected edge filter is proposed to smooth the initial sparse blur map based on pixel connectivity within detected edge contours. Then a fast-guided filter is used to propagate the sparse blur map through the whole image.

Region-based methods directly estimate the defocus amount from the local patches centered at the current pixel \cite{trouve2011single,shi2015just}. For each local patch in the image, Trouvé et al. \cite{trouve2011single} use a maximum likelihood method to select the local blur from a set of PSF candidates. 
In \cite{d2016non}, a machine learning approach based on the regression tree fields is used to train a model able to regress a coherent defocus blur map of the image, labeling each pixel by the scale of a defocus point spread function. 
Shi et al. \cite{shi2015just} propose a method based on dictionary learning using sharp and slightly blurred patches.

Recently, CNN-based methods have been proposed for defocus map estimation. Yan and Shao \cite{yan2016blind} propose a method that first classifies the blur type and then estimates the blur parameter using a general regression neural network. Zhao et al. \cite{zhao2018defocus} propose a method that detects defocus blur by using a bottom-top-bottom fully convolutional network. Since defocus blur detection only classifies each pixel as blur pixel and the non-blur pixel while the defocus map estimation needs to estimate the blue level of each pixel
so the defocus blur detection can be considered as the loose formulation of the defocus map estimation.
Lee et al. \cite{lee2019deep} propose an end-to-end CNN-based method (DME-Net) for spatially varying defocus map estimation, for which they also created a synthetic dataset. They employ a domain adaptation method to address the gap between real and synthetic datasets. Theirs is the first truly CNN-based defocus map estimation method. Similar to \cite{lee2019deep}, Ma et al., \cite{ma2021defocus} train their model in an end-to-end manner for defocus map estimation. These methods suffer from the lack of sufficient real data.

\subsection{Non-Blind Defocus Deblurring}

Non-blind defocus deblurring, which assumes either the defocus map or the blur kernel to be known,  is an ill-posed problem since some information will inevitably be lost during the blurring process. Most classical methods impose some image priors to regularize the solution, for example, patch-based prior \cite{zoran2011learning}, hyper-Laplacian prior \cite{krishnan2009fast}, and local color prior \cite{joshi2009image}. These methods usually need computationally expensive iterative optimization. Recently, researchers have shown that CNN-based methods will be better than these classical approaches in terms of accuracy and efficiency \cite{schuler2013machine,xu2014deep,zhang2017learning}. However, most existing methods, such as Schuler et al. \cite{schuler2013machine} and Xu et al. \cite{xu2014deep}, need to be trained for each specific blur kernel;  when we encounter images with unseen kernels, we need to retrain the networks. 

\subsection{Single Image Defocus Deblurring}

Many studies have been conducted for single-image defocus deblurring. Conventional methods typically decompose the problem into two steps. The first step is to estimate the defocus map of an input image, which indicates the blur level for each pixel \cite{tai2009single,zhuo2011defocus,karaali2017edge}. The second step is the non-blind deconvolution \cite{krishnan2009fast,levin2007image}, where the defocus map estimated in the first step is used. Most of the methods employing this two-step approach focus on improving the accuracy of the first step, i.e., the defocus map estimation, since the small error on the defocus map will significantly deteriorate the final deblurring performance \cite{karaali2017edge,park2017unified,shi2015just,d2016non}. 

% {\color{red} [Is this paragraph for defocus map estimation? If so, we have already explained it. \cite{lee2019deep} and \cite{ma2021defocus} are mentioned earlier. Needs some reorganization.]}
% A simple but effective method to estimate the defocus map is to compute a sparse map on image edges and then propagate this sparse map across the input image to achieve the final defocus map. Shi et al. \cite{shi2015just} detect just noticeable blur via sparse representation and image decomposition. In \cite{karaali2017edge}, a connected edge filter is proposed to smooth the initial sparse blur map based on pixel connectivity within detected edge contours. Then a fast-guided filter is used to propagate the sparse blur map through the whole image. 

% Another strategy for defocus map estimation is to utilize machine learning methods. In \cite{d2016non}, a machine learning approach based on the regression tree fields is used to train a model able to regress a coherent defocus blur map of the image, labeling each pixel by the scale of a defocus point spread function. 
% Lee et al. \cite{lee2019deep} estimate defocus maps using a domain adaption strategy in deep learning. Following the conventional two-step strategy,  after estimating the defocus map using a convolutional neural network, Ma et al. \cite{ma2021defocus} concatenates the defocus map and the defocus image as input of the neural network to achieve all sharp images. 

Another approach is to train a network to directly predict the deblurred image from a blurry input image in an end-to-end fashion. Abuolaim et al. \cite{abuolaim2020defocus} use a network with the UNet structure to do this. Lee et al. \cite{lee2021iterative} propose an iterative filter adaptive module to handle spatially-varying and large defocus blur. They also propose a training scheme based on defocus disparity estimation and re-blurring to boost deblurring quality. Son et al. \cite{son2021single} propose to simulate inverse kernels by kernel sharing parallel atrous convolution block. However, these methods based on direct prediction depend too much on the training data, and we believe their performance is suboptimal.

%------------------------------------------------------------------------- 

\begin{figure}[tb]
    \centering
    \includegraphics[width=1.0\linewidth]{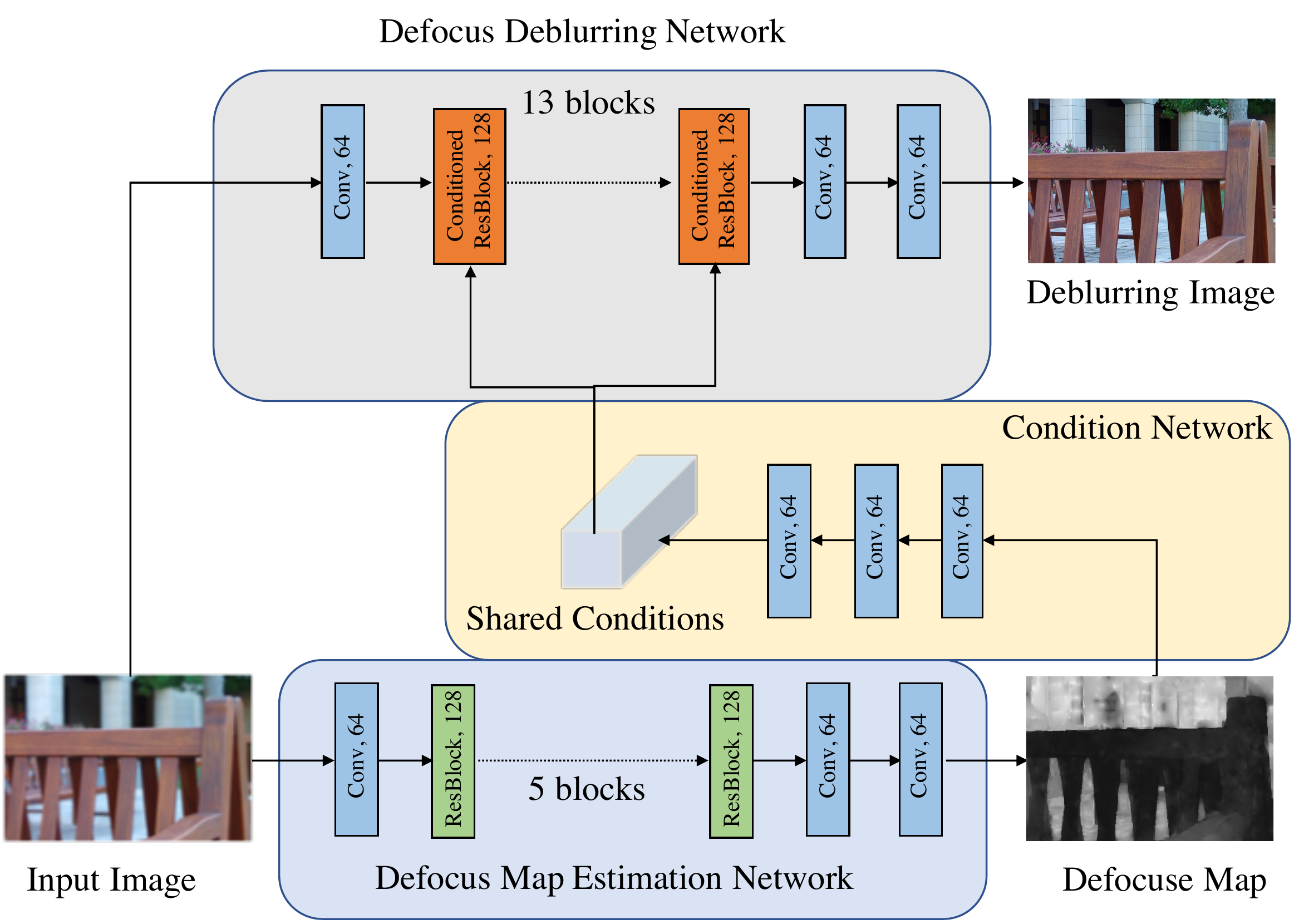}
    \caption{Architecture of the proposed network.}
    \label{fig:cp4_arch}
\end{figure}

\section{Proposed Method}
We propose a network to remove spatial variant defocus blur from a single blurry image. Inspired by the previous deblurring methods that decompose the blind deblurring task into defocus map estimation task and non-blind deblurring \cite{tai2009single,zhuo2011defocus,karaali2017edge,ma2021defocus}. Our network also first estimates the defocus map as an intermediate result. For CNN-based based methods, they directly use the feed-forward networks to learn the mapping from the input blurring images, while the previous CNN-based two steps method directly concatenates the defocus map with the input blurring images as the input as the non-blind deblurring network. The defocus map, i.e., knowing the blur level for each pixel, is very important information for the non-blind deblurring. But previous method utilizes the defocus map by simple concatenation. We think this leads to suboptimal results. 

Considering the spatial variant property of the defocus blur and the blur level indicated in the defocus map, we employ the defocus map as conditional guidance to adjust the features from the input blurring images instead of simple concatenation.
For better usage of the defocus map, we employ an affine transform to adjust the features from the input blurring images for each pixel. By considering the same blur level on regions in blurring image, we further decompose the parameters for scaling in the affine transform into spatial and channel dimensions to reduce the redundancy and improve the capacity. 

Figure~\ref{fig:cp4_arch} shows our network which consists of three sub-network: defocus map estimation network, the condition network and the deblurring network. 

% Given a single blurry image $I_b$, the defocus deblurring aims at reconstrcuting an all-sharp image. 
\begin{figure}[htbp]
    \centering
    \includegraphics[width=.7\linewidth]{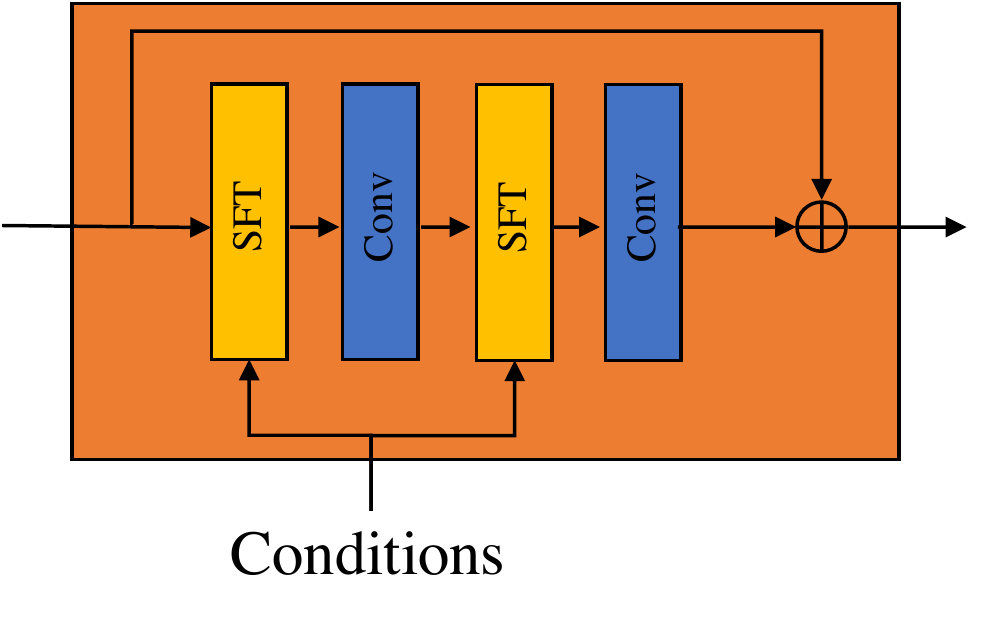}
    \caption{Architecture of the conditioned res-block.}
    \label{fig:cp4_con_res}
\end{figure}

\subsection{Defocus Map Estimation Network}
The defocus map estimation network utilizes several residual blocks to ease the training procedure and maximize the information flow, which takes the blurry images as input and then estimates the defocus map.  

\subsection{Condition Network}
The condition network only consists of three convolutional layers, which takes the estimated defocus map as input and maps it into the feature space as conditions that are afterward used to modulate the intermediate features in the defocus deblurring network. The key to reconstructing the all-sharp images is to recover the missing details in the out-of-focus regions in the input blurrying images. Different areas in one image have different contents and blur levels. Further, different images also have different contents and blur levels. Therefore, it is necessary to deal with input images with location-specific and image-specific operations. However, the convolutional layer is spatial invariant. Inspired by spatial feature transform (SFT) \cite{wang2018recovering}, we introduce a network with SFT to perform a spatial variant adjustment. Here the condition network is to generate the conditions for SFT from the defocus map. 

\begin{figure}[htbp]
    \centering
    \includegraphics[width=.7\linewidth]{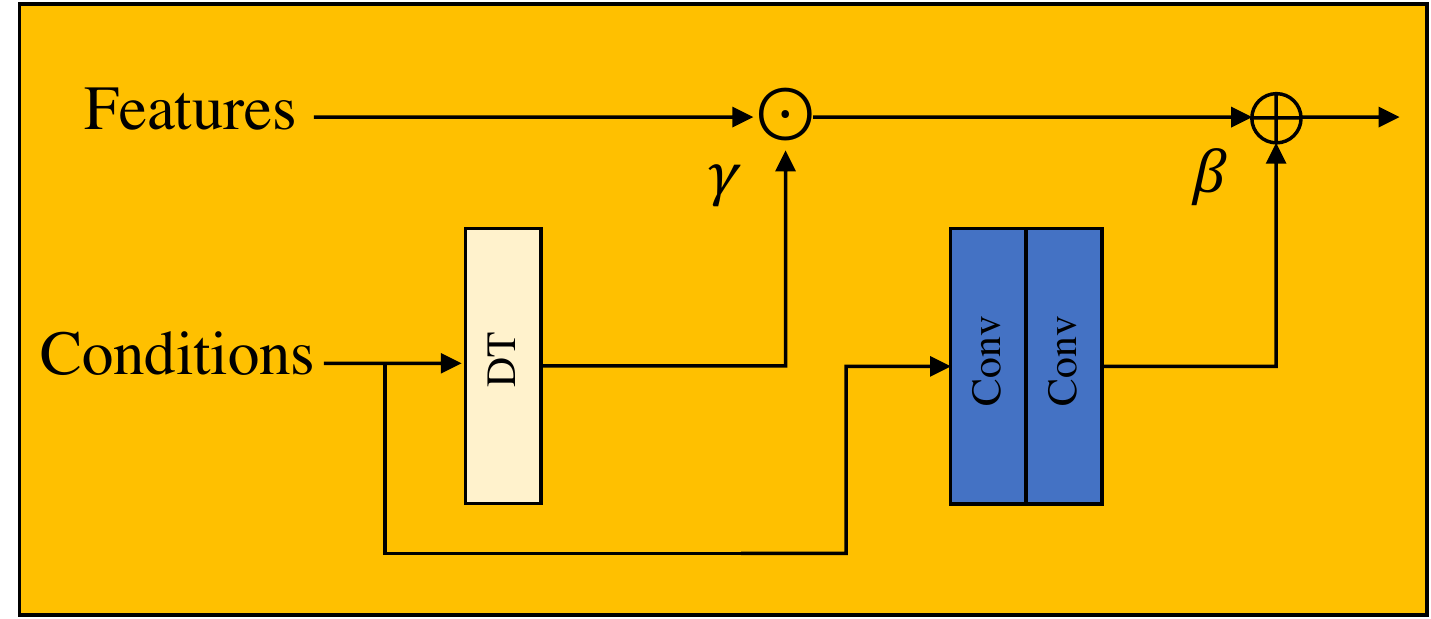}
    \caption{Architecture of the feature transform.}
    \label{fig:cp4_sft}
\end{figure}

\subsection{Defocus Deblurring Network}
Defocus deblurring network takes not only the blurring image but also the conditions from the condition network as input. There are some conditioned residual blocks in the defocus deblurring network and the Figure~\ref{fig:cp4_con_res} shows the details of conditioned residual blocks. The key component in conditioned residual blocks is the SFT layer. The structure of SFT layer is shown in Figure~\ref{fig:cp4_sft}. The SFT learns a mapping function that generates the paired modulation parameters $\gamma$ and $\beta$ based on the defocus map as prior. The learned parameters adaptively adjust the output by an affine transform for each pixel to the intermediate features on the defocus deblurring network. Specifically, the SFT layer can be described as,

\begin{equation}
SFT(F) = \gamma \odot F + \beta
\end{equation}

\noindent $F \in \mathbb{R}^{C\times H\times W}$ is the intermediate features in defocus deblurring network and $\gamma \in \mathbb{R}^{C \times H\times W}$ and $\beta \in \mathbb{R}^{C \times H\times W}$ are the parameters for modulation. $\odot$ is the element-wise multiplication.

\begin{figure}[htbp]
    \centering
    \includegraphics[width=.8\linewidth]{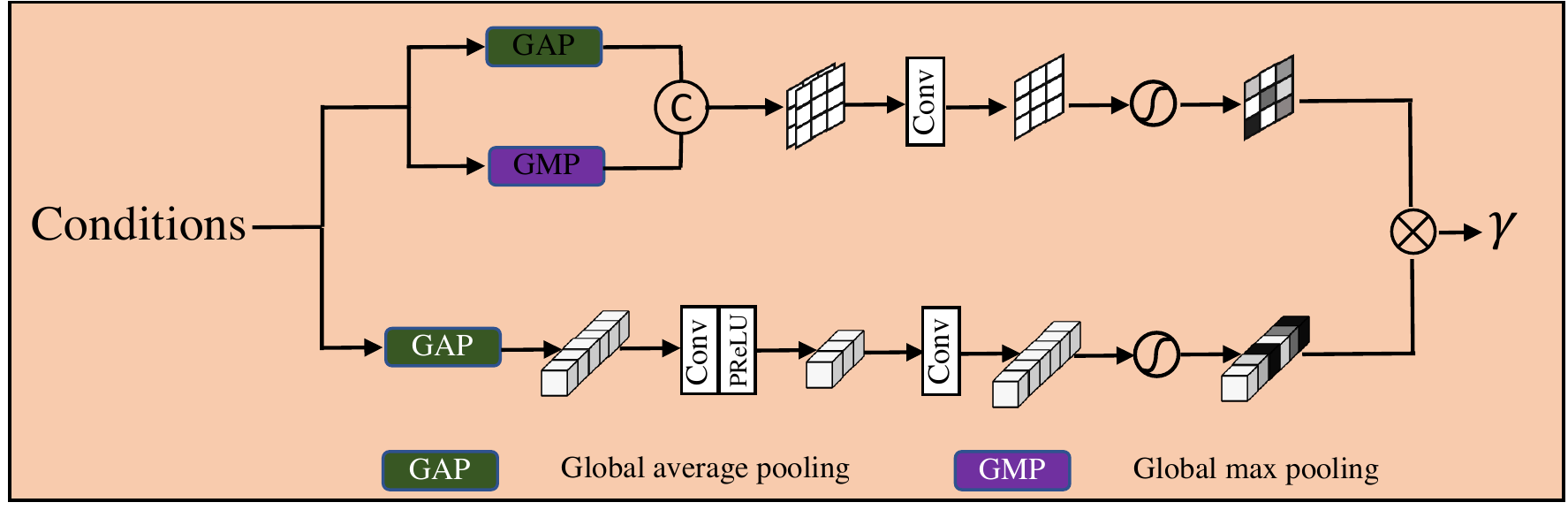}
    \caption{Architecture of the decomposition transform.}
    \label{fig:cp4_dt}
\end{figure}

Further, as shown in Figure~\ref{fig:cp4_arch}, there are some regions with the same value in defocus map causing the redundancy on the conditions. Inspired by channel attention \cite{hu2018squeeze} and spatial attention \cite{woo2018cbam}, we perform decomposition on the generation for $\gamma$ in the channel dimension $\mathbb{R}^{C \times 1\times 1}$ and spatial dimension $\mathbb{R}^{1 \times H\times W}$ as shown in Figure~\ref{fig:cp4_dt}. 
It should be noted that we do not apply the decomposition transform to the generation of $\beta$ since there are differences for each pixel value, even though their blur levels are the same. Thus, the $\beta$ without decomposition transform can provide the detailed compensation for the deblurring network. By using this modulation strategy, our method can better use the important information in defocus map. 

%------------------------------------------------------------------------- 
\subsection{Loss Function}
The proposed network consists of three sub-network: defocus map estimation network, condition network and the deblurring network.  Following the previous work \cite{ma2021defocus}, we use the $L_1$ norm loss and $L_2$ norm loss for training. Specifically, for defocus map estimation, the loss is computed as follows,

\begin{equation}
  L_{dme} = ||DM_{e} - DM_{gt}||_1
\end{equation}
\noindent where $DM_e$ is the estimated defocus map and the $DM_{gt}$ is the ground truth for the defocus map.

While for the defocus deblurring, we exploit two losses: $L_{db}$ and $L_{wd}$. $L_{db}$ is defined as 

\begin{equation}
    L_{df} = ||I_{df} - I_{gt}||_1
\end{equation}
while $L_{wd}$ is defined as
\begin{equation}
    L_{wd} = ||W_{dm} * (I_{df} - I_{gt})||_2
\end{equation}
\noindent where $I_{df}$ is the deblurring result of the deblurring network and the $I_{gt}$ is the all-sharp ground truth image. $W_{dm}$ is defined as
\begin{equation}
    W_{dm} = \frac{DM_{gt}}{mean(DM_{gt})}
\end{equation}

Similar to \cite{ma2021defocus}, we train our network for three stages. In stage one and stage two, the two networks are jointly trained for 400 and 200 epochs, respectively. For stage one, the ground truth defocus
map is set as the input of the defocus deblurring network, which avoids divergence caused by random output of the defocus map
estimation network. For stage two, the output of the defocus map estimation network is set as the input of the defocus deblurring network to jointly training the whole network. For both stage one and stage two, the loss used for training is

\begin{equation}
    Loss_1 = \lambda_1 \times L_{dme} + \lambda_2 \times L_{df}
\end{equation}
In stage three, we use the following loss to finetune the network for another 400 epochs,
\begin{equation}
    Loss_2 = \lambda_2 \times L_{df} + \lambda_3 \times L_{wd}
\end{equation}

In this work, we set the weights for loss function as $\lambda_1=0.2$, $\lambda_2=0.9$ and $\lambda_3=0.1$

%------------------------------------------------------------------------- 

\section{Experiments}

\subsection{Experimental Settings}
\subsubsection{Training Data}
To train our network, we use the defocus image deblurring dataset in \cite{ma2021defocus} which consists of both the defocus map ground truth and all-sharp image ground truth.

\subsubsection{Test Data}
Followed the previous work \cite{ma2021defocus}, we evaluate the proposed method on the Realistic
dataset \cite{d2016non} and the DED test dataset \cite{ma2021defocus}.

\subsubsection{Implementation Details}
For training, we use the Adam optimizer \cite{KingmaB14} with $\beta_1=0.9$, $\beta_2=0.999$ with initial learning rate $1\times10^{-4}$ and set the batch size to 16. And the number of epochs for each stage is mentioned above. The input blurring image, the defocus map ground truth and the corresponding all-sharp ground truth are randomly cropped into patches with 256$\times$256. We also apply other data augmentation like random rotation and flipping to avoid over-fitting. We implement our model by PyTorch \cite{paszke2017automatic} platform and train our model on NVIDIA GeForce RTX 2080 GPUs.

%------------------------------------------------------------------------- 

\subsection{Experimental Resutls}
We first evaluate the performance of defocus map estimation. We compare the proposed method with the methods of Karaali and Jung \cite{karaali2017edge} and the recent deep learning-based DME-Net \cite{lee2019deep} and DID-ANet \cite{ma2021defocus} for defoucs map estimation. Mean absolute error (MAE) and mean squared error (MSE) are used as the evaluation metrics. Table~\ref{Table:cp4_main_map} shows the quantitative results on Realistic dataset. The proposed method achieves almost the same performance with DID-ANet \cite{ma2021defocus} since we adopt the same network with DID-ANet \cite{ma2021defocus} and is comparable with DMENet \cite{lee2019deep} on Realistic dataset.  

% Several samples for visualization of defocus map estimation are shown in Figure~\ref{fig:map_real} and \ref{fig:map_ded}. 

We compare the proposed method with the DME-Net \cite{lee2019deep} that estimates the defocus map by CNN and achieves deblurring by conventional deconvolution \cite{krishnan2009fast}, two CNN-based methods for defocus deblurring (DPDDNet \cite{abuolaim2020defocus} and IFAN \cite{lee2021iterative}), three CNN-based state-of-arts methods (MPRNet \cite{zamir2021multi}, MIMO-UNet \cite{cho2021rethinking} and NAFT \cite{chen2022simple}) for motion deblurring and two-steps CNN-based method (DID-ANet \cite{ma2021defocus}). All the methods are fine-tuned on the training set of DED dataset. And the training settings are set separately based on the original paper. The epochs for fine-tuning are set to 600 to ensure the models are convergent. 

\begin{table}[htbp]
\begin{center}
%   \vspace*{-4mm}
\caption{Quantitative Results for Defocus Map Estimation}
\label{Table:cp4_main_map} \small
\setlength{\tabcolsep}{5pt}
\begin{tabular}{l|*{10}{c}}
\hline 
  Methods                     & MAE, Realistic  & MSE, Realistic \\%&  PSNR, DED-test &SSIM, DED-test \\
\hline 
Karraali \cite{karaali2017edge} & 0.3102 & 0.1245 \\
DMENet \cite{lee2019deep}  &  \textbf{0.1191} & \textbf{0.0242} \\
DID-ANet \cite{ma2021defocus} & 0.1331 & 0.0312 \\
Ours & 0.1303 &  0.0299 \\

\hline
\end{tabular}
\end{center}
\end{table}

We use the Peak Signal to Noise Ratio (PSNR) and Structural Similarity (SSIM) for evaluation metrics. The quantitative
results are shown in Table \ref{Table:cp4_main} where the best results are in bold. For Realistic dataset, %\footnote{The ground truth for DED dataset is unavailable now. Here we only use the Realistic dataset for quantitative comparisons.}, 
the proposed method achieves better performance than the other methods in terms of PSNR and SSIM. The CNN-based methods for motion blur and defocus blur do not achieve good performance compared with the two-steps methods (e.g., DID-ANet \cite{ma2021defocus}), since they do not utilize the information from the defocus map which is essential for defocus image blurring. Although The proposed method achieves almost the same performance with DID-ANet \cite{ma2021defocus} in the defoucs map estimation, it still achieves much higher performance than DID-ANet \cite{ma2021defocus} with more than 0.6db improvement, which shows the effectiveness of the proposed method. And also compared DME-Net \cite{lee2019deep} with conventional deconvolution \cite{krishnan2009fast}, the CNN-based methods (e.g., DID-ANet \cite{ma2021defocus}) can achieve much better performance. 
% For both Realistic dataset and the DED-test dataset, the proposed method achieves better performance than the other methods in terms of PSNR and SSIM. 

\begin{table}[htbp]
\begin{center}
%   \vspace*{-4mm}
\caption{Quantitative Results for Defocus Image Deblurring}
\label{Table:cp4_main} \small
\setlength{\tabcolsep}{5pt}
\begin{tabular}{l|*{10}{c}}
\hline 
                       & PSNR, Realistic  & SSIM, Realistic \\%&  PSNR, DED-test &SSIM, DED-test \\
\hline 
DMENet \cite{lee2019deep}  & 24.0397 & 0.7180 \\ % & & \\	
DPDDNet \cite{abuolaim2020defocus} & 24.7232 & 0.7640 \\ % & & \\
IFAN \cite{lee2021iterative}& 24.9261  & 0.8211 \\ %& & \\
MPRNet \cite{zamir2021multi}& 25.2738 &  0.7869\\ % & & \\	
MIMO-UNet \cite{cho2021rethinking}& 25.3430 &  0.7962\\ % & & \\	
NAFT \cite{chen2022simple}& 24.3461 &  0.7484\\ % & & \\	
DID-ANet \cite{ma2021defocus} & 25.9491 & 0.8204\\ % & &\\ 	
Ours &  \textbf{26.5535} & \textbf{0.8465} \\ %& & \\

\hline
\end{tabular}
\end{center}
\end{table}

\begin{figure}[htbp]
    \centering
    \includegraphics[width=1\linewidth]{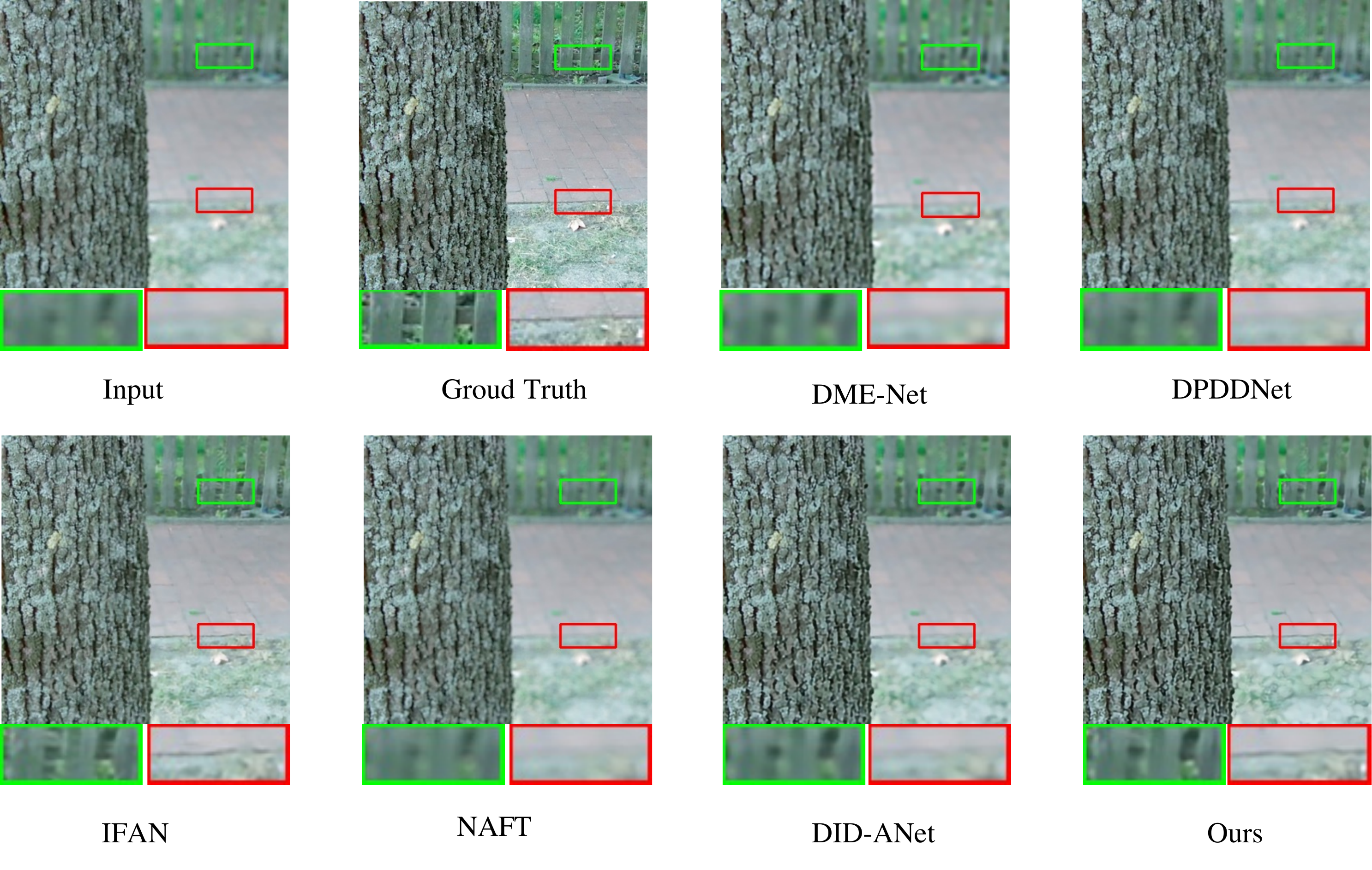}
    \caption{Visual comparison of defocus image deblurring on realistic (08).}
    \label{fig:rtf08}
\end{figure}

\begin{figure}[htbp]
    \centering
    \includegraphics[width=1\linewidth]{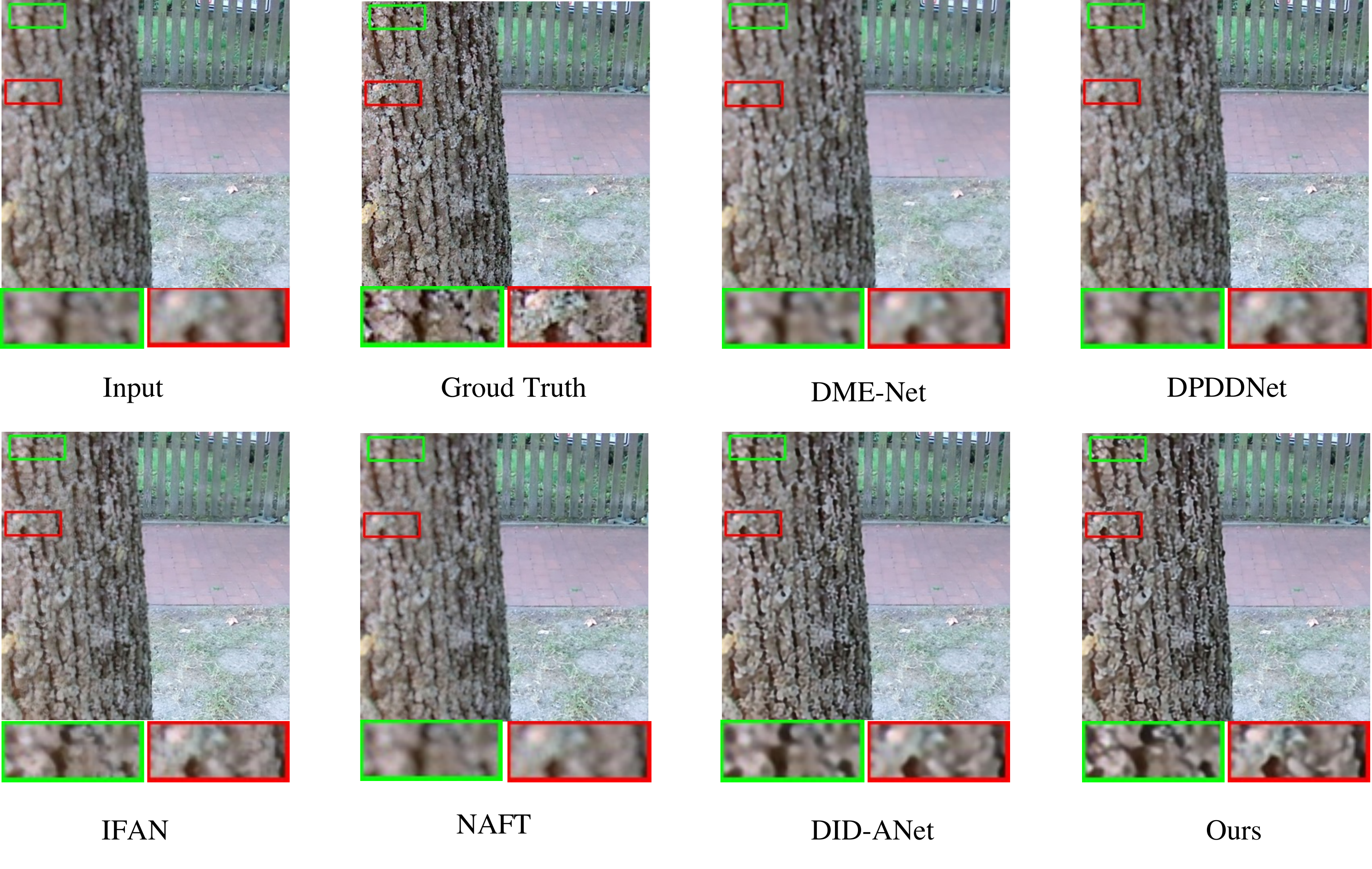}
    \caption{Visual comparison of defocus image deblurring on realistic (09).}
    \label{fig:rtf09}
\end{figure}

\begin{figure}[htbp]
    \centering
    \includegraphics[width=1\linewidth]{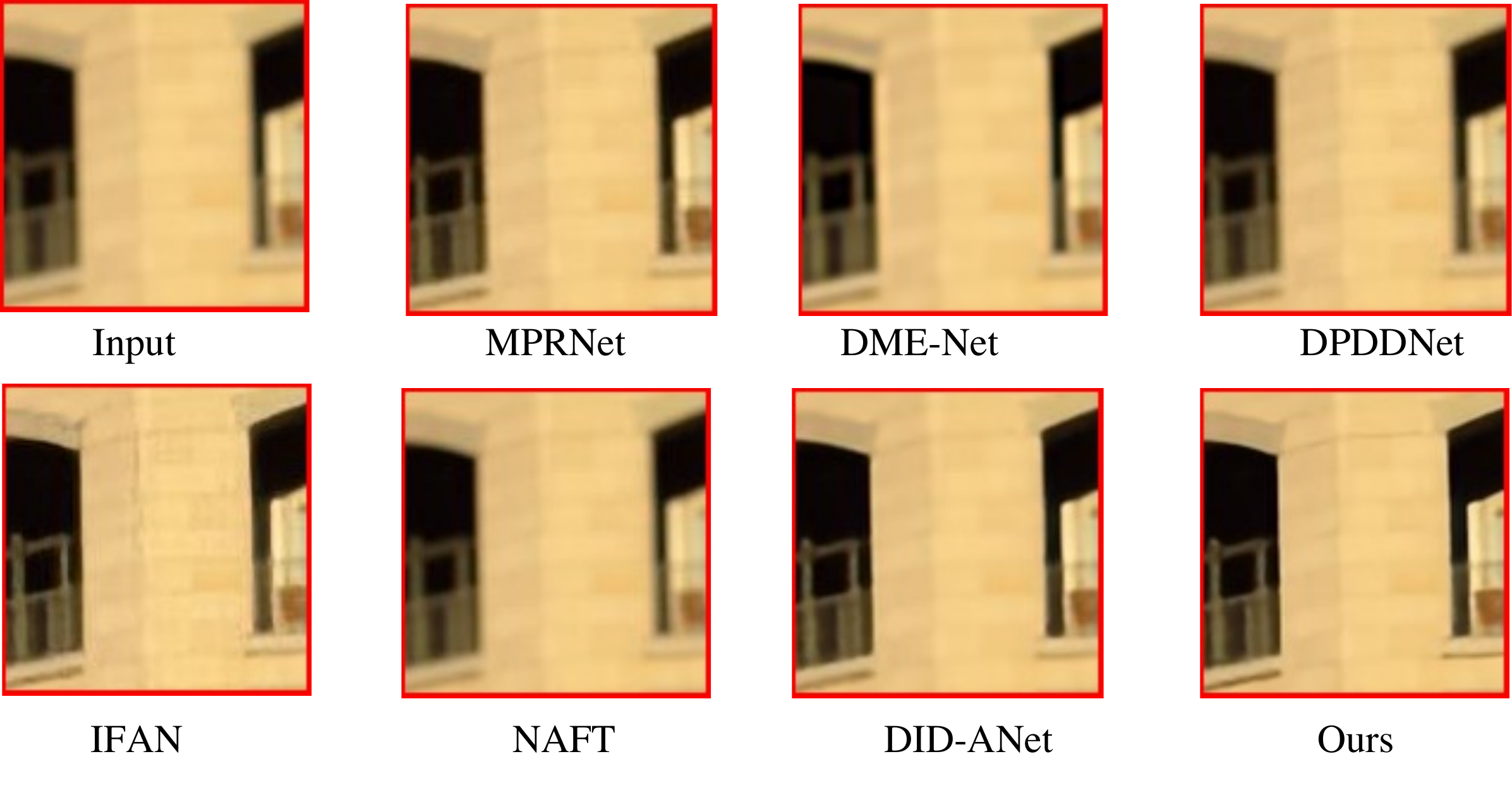}
    \caption{Visual comparison of defocus image deblurring on DED dataset (604).}
    \label{fig:ded_604}
\end{figure}

\begin{figure}[htbp]
    \centering
    \includegraphics[width=1\linewidth]{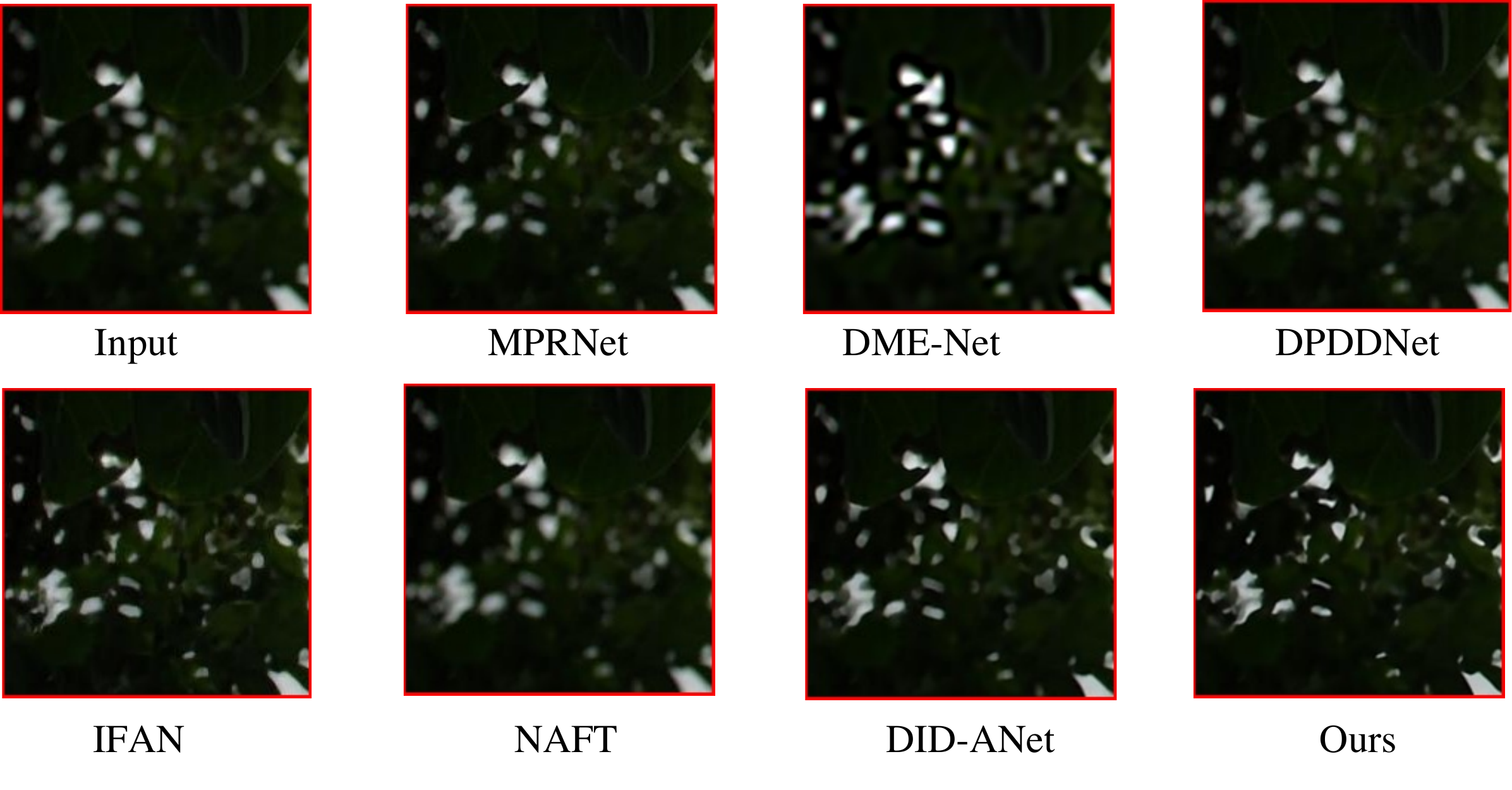}
    \caption{Visual comparison of defocus image deblurring on DED dataset (794).}
    \label{fig:ded_794}
\end{figure}

\begin{figure}[htbp]
    \centering
    \includegraphics[width=1\linewidth]{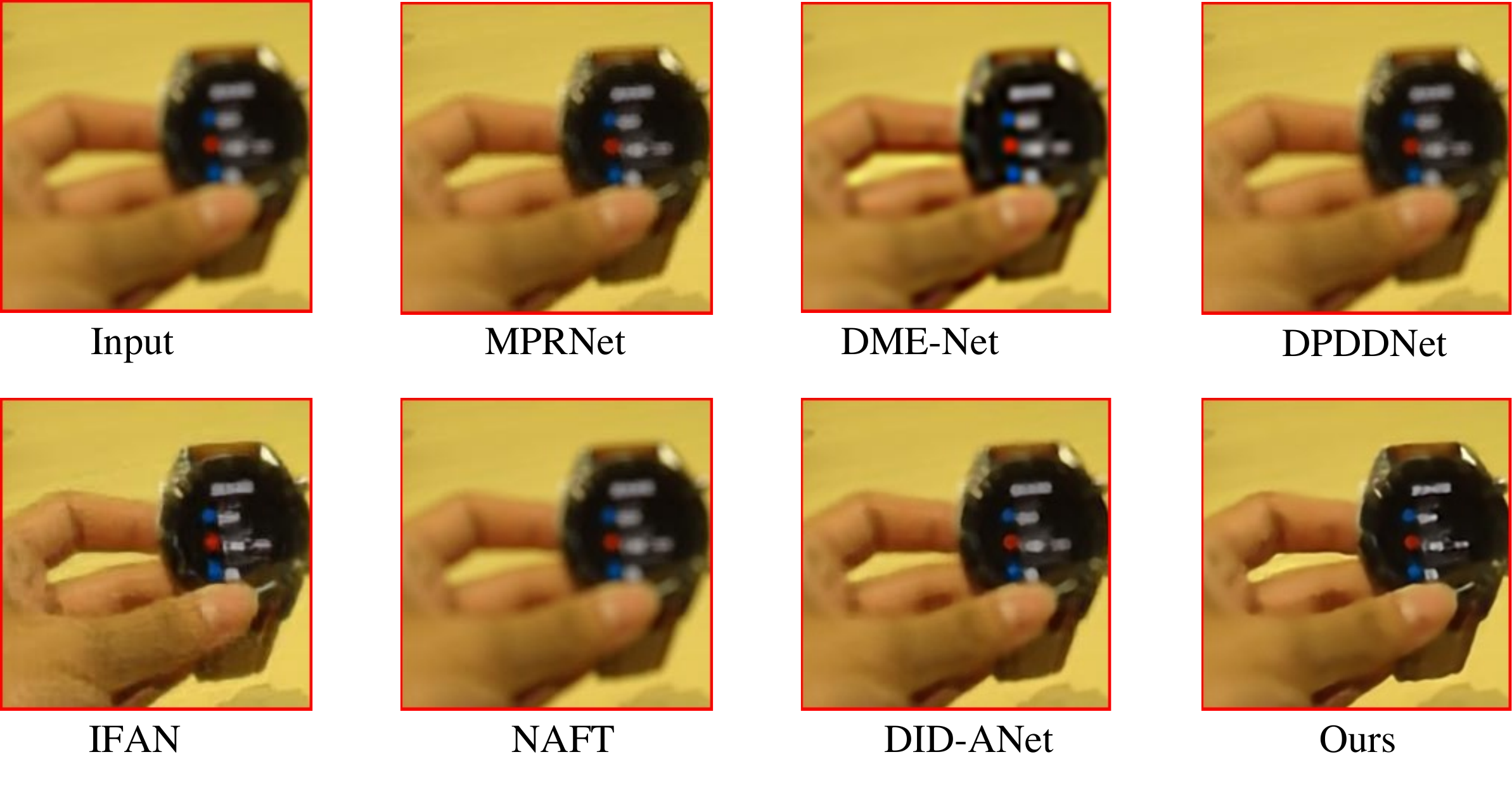}
    \caption{Visual comparison of defocus image deblurring on DED dataset (971).}
    \label{fig:ded_971}
\end{figure}

Several visualization results are shown in Figure~\ref{fig:rtf08}-\ref{fig:ded_971}. Among them, Figure~\ref{fig:rtf08} and ~\ref{fig:rtf09} are from Realistic test set while Figure~\ref{fig:ded_604}-\ref{fig:ded_971} are from DED test set. Since the image size for the DED test set is large, we crop the images from DED test set and zoom them for a better view. 

The images from Figure~\ref{fig:rtf08} and ~\ref{fig:rtf09} are almost the same contents but have different focal plane. The image (08) in the Figure~\ref{fig:rtf08} focuses on the tree then the background is blurry. As shown in Figure~\ref{fig:rtf08}, the structures on the road and the fence are clearer in our result. While in Figure~\ref{fig:rtf09}, ours results have clearer texture compared with other methods. As shown in Figure~\ref{fig:ded_604}, our method can achieve the sharper edge on the wall with fewer artifacts. The light region in input image shown in Figure~\ref{fig:ded_794} is blurry, while our method can successfully recover the sharp shape of the light region. The hand and the watch are more realistic and clearer in our result.  

In contrast, the previous works cannot address the defocus blur very well. The DME-Net \cite{lee2019deep}  estimates the defocus map and then uses the deconvolution deblurring \cite{krishnan2009fast} to achieve defocus deblurring. Their results look blurry compared with ours. On the other hand, the CNN-based based methods (DPDDNet \cite{abuolaim2020defocus} and IFAN \cite{lee2021iterative}) also achieve blurry results though the model are finetuned on the DED dataset. For the state-of-the-art method in motion deblurring methods, NAFT \cite{chen2022simple} cannot deal with the defocus blur very well. For the two-steps methods, DID-ANet achieves better performance than the above methods. However, the important information in the defocus map can not be fully utilized by simple concatenation. On the contrary, the proposed method exploits the information from the defocus map by the SFT with the decomposition method, which leads to higher PSNR and SSIM than previous methods, as shown in Table \ref{Table:cp4_main}

\begin{table}[htbp]
\begin{center}
%   \vspace*{-4mm}
\caption{Ablation Study on Results for Defocus Image Deblurring}
\label{Table:cp4_ablation} \small
\setlength{\tabcolsep}{5pt}
\begin{tabular}{l|*{10}{c}}
\hline 
    Methods                   & PSNR, Realistic  & SSIM, Realistic \\ %&  PSNR, DED-test &SSIM, DED-test \\
\hline 
baseline-S1 & 25.6443 & 0.8077\\ % & & \\	
+SFT-S1 & 26.1286 & 0.8294 \\ %& & \\
+SFT-Dec-S1 & 26.0837  & 0.8289 \\ % & & \\
+SFT-FDec-S1 & 25.9939 & 0.8253 \\ %\\
% +SFT-Dec-DME-S1 & 26.4595 & 0.8402 \\ %\\

\hline
baseline-S2 & 25.7936 & 0.8118 \\ % & & \\	
+SFT-S2 & 26.2533 & 0.8307 \\ %& & \\
+SFT-Dec-S2 & 26.4275  & 0.8310\\ % & & \\
+SFT-FDec-S2 & 26.1735 & 0.8307\\ %\\
% +SFT-Dec-DME-S2 & \textbf{26.6031} & \textbf{0.8482} \\ %\\

\hline
baseline-S3 & 25.9638 & 0.8191 \\ %& & \\	
+SFT-S3 & 26.3973 & 0.8379 \\ % & & \\
+SFT-Dec-S3 & \textbf{26.5535} & \textbf{0.8465} \\ % & & \\
+SFT-FDec-S3 & 26.4327 & 0.8400 \\ %\\
% +SFT-Dec-DME-S3 & \textbf{26.6129} & 0.8455 \\ %\\

\hline
+SFT-Dec-end & 26.1143 & 0.8259\\ %\\

\hline
\end{tabular}
\end{center}
\end{table}

\subsection{Ablation Study}
We demonstrate the effectiveness of each component in the proposed method on the Realistic dataset. The results are shown in Table \ref{Table:cp4_ablation}. For the baseline model, we remove the SFT with the decomposition method. For the "baseline+SFT", it means that we use the original SFT in the baseline. For the "baseline+SFT-Dec", it means that we use the SFT with the decomposition on $\gamma$ into the baseline. For the "baseline+SFT-FDec", it means that we use the SFT with the decomposition on both $\gamma$ and $\beta$ into the baseline. S1, S2 and S3 indicate the different training stages, respectively while "-end" means the model is trained in an end-to-end manner with the supervision of the defocus map and the all sharp ground truth. 

As shown in Table~\ref{Table:cp4_ablation}, for all these methods, the performance increases after each stage training. While the final model "baseline+SFT-Dec-s3" achieves the best performance compared with other ablated networks. At the same time, if the final model is trained in an end-to-end manner, the performance decreases by a large margin. 
By introducing the SFT, the performance increases about 0.4db in PSNR and 0.02 in SSIM compared with the baseline model. While incorporating the decomposition method, the performance increases by about 0.6db in PSNR and 0.03 in SSIM compared with the baseline model. However, if we also apply the decomposition method to the $\beta$, the performance drops about 0.1db in PSNR. This is because the $\beta$ provides the detailed information for modulation but the decomposition method will remove these details to some extent.

%------------------------------------------------------------------------- 

\section{Conclusion}

In this paper, we propose a new method for defocus image deblurring by first defocus map estimation and then the defocus deblurring direction. The defocus map is important information for defocus image deblurring since it contains the blur level for each pixel. To remove the spatial variant blur, we introduce the spatial feature transform and the decomposition technique to perform spatial modulation based on the defocus map. The experimental results have validated the effectiveness of the proposed approach. 

%------------------------------------------------------------------------- 

%%%%%%%%% REFERENCES
{\small
\bibliographystyle{ieee_fullname}
\bibliography{egbib}
}

\end{document}